\documentclass[10pt,twocolumn,letterpaper]{article}

\usepackage[pagenumbers]{cvpr}      % To produce the REVIEW version
\usepackage{amsmath,amssymb,amsfonts,mathtools}
\usepackage{amsthm}
\usepackage{booktabs}
\usepackage{algorithm}
\usepackage{algorithmic}
\usepackage{graphicx}
\usepackage{microtype}
\usepackage{tikz}
\usetikzlibrary{positioning,arrows.meta}

\definecolor{cvprblue}{rgb}{0.21,0.49,0.74}
\usepackage[pagebackref,breaklinks,colorlinks,allcolors=cvprblue]{hyperref}

\theoremstyle{definition}

\newcommand{\E}{\mathbb{E}}
\newcommand{\norm}[1]{\left\lVert #1 \right\rVert}

\newcommand{\defeq}{\mathrel{\mathop:}=}

\def\paperID{0007}
\def\confName{CVPR}
\def\confYear{2026}

\begin{document}

\title{Sharpness-Aware Surrogate Training for\\on-sensor Spiking Neural Networks}

\author{%
Maximilian Nicholson\\
University of Bath, United Kingdom\\
{\tt\small mn866@bath.ac.uk}
}

\maketitle

\begin{abstract}
Spiking neural networks (SNNs) are a natural computational model for on-sensor and near-sensor vision, where event driven processors must operate under strict power budgets with hard binary spikes. However, models trained with surrogate gradients often degrade sharply when the smooth surrogate nonlinearity is replaced by a hard threshold at deployment; a surrogate-to-hard transfer gap that directly limits on-sensor accuracy. We study Sharpness-Aware Surrogate Training (SAST), which applies Sharpness-Aware Minimization (SAM) to a surrogate-forward SNN so that the training objective is smooth and the gradient is exact, and position it as one gap-reduction strategy under the tested settings rather than the only viable mechanism. Under explicit contraction assumptions we provide state-stability, input-Lipschitz, and smoothness bounds, together with a corresponding nonconvex convergence result. On two event-camera benchmarks, swap-only hard-spike accuracy improves from 65.7\% to 94.7\% on N-MNIST and from 31.8\% to 63.3\% on DVS Gesture. Under a hardware-aware inference simulation (INT8/INT4 weight quantization, fixed-point membrane potentials, discrete leak factors), SAST remains strong: on N-MNIST, hard-spike accuracy improves from 47.6\% to 96.9\% (INT8) and from 43.2\% to 81.0\% (INT4), while on DVS Gesture it improves from 25.3\% to 47.6\% (INT8) and from 26.0\% to 43.8\% (INT4). SynOps also decrease under the same hardware-aware setting, including 1734k$\rightarrow$1315k (N-MNIST, INT8) and 86221k$\rightarrow$4323k (DVS Gesture, INT8). These results suggest that SAST is a promising component in a broader toolbox for on-sensor spiking inference under the tested settings.
\end{abstract}

%------------------------------------------------------------------------
\section{Introduction}
\label{sec:intro}

On-sensor vision aims to unify sensing and computation on a single chip, eliminating costly data transfers and enabling real-time, sub-watt operation~\cite{lichtsteiner2008dvs,gallego2022eventvision,davies2018loihi,merolla2014truenorth}.
Event-based sensors such as dynamic vision sensors (DVS) produce temporally sparse, asynchronous streams that align naturally with spiking neural networks (SNNs)~\cite{maass1997networks,indiveri2015memory,roy2019towards}.
Because spikes are binary and event-driven, SNNs are well suited to the massively parallel, low-power processing arrays---whether neuromorphic, PPA-based, or analog---that define the on-sensor paradigm. The central training challenge is that the spike function is discontinuous. Surrogate-gradient methods replace its derivative with a smooth proxy during backpropagation~\cite{neftci_surrogate,zenke_superspike,bellec2018l2l,shrestha2018slayer,wu2018stbp,eshraghian2023lessons}, but a persistent gap remains between the surrogate model used during training and the hard-threshold model required at deployment. When many membrane potentials cluster near threshold, the smooth surrogate emits graded activations while the on-sensor hardware must commit to $0$ or $1$; this mismatch compounds across time steps and layers, degrading accuracy at inference. For on-sensor deployment, where the final model \emph{must} use hard spikes, this transfer gap is a key obstacle. Sharpness-Aware Minimization (SAM)~\cite{foret_sam,kwon2021asam} optimizes a neighborhood worst-case loss and improves generalization in conventional networks. We apply SAM not to a hard-forward/surrogate-backward estimator, but to a \emph{surrogate-forward} SNN whose dynamics already use a smooth spike approximation. This makes the training objective genuinely smooth, so backpropagation through time computes the exact gradient and the analysis applies directly to the model being optimized.
We call the resulting method \emph{Sharpness-Aware Surrogate Training} (SAST).

\textbf{Contributions.}
We present SAST as a training-time method for improving on-sensor deployability under hard-spike inference with quantized weights and fixed-point membrane constraints, emphasizing cross-setting performance under the tested settings rather than method exclusivity.
(i) We formalize SAST for multi-layer LIF SNNs with state-stability, input-Lipschitz, smoothness, and convergence results under explicit assumptions.
(ii) On N-MNIST~\cite{nmnist} and DVS Gesture~\cite{amir2017lowpower} with a small fully-connected SNN budgeted at ${\sim}0.40$M parameters (implemented as $2312\!\rightarrow\!168\!\rightarrow\!64\!\rightarrow\!10$ for N-MNIST and $4608\!\rightarrow\!80\!\rightarrow\!336\!\rightarrow\!11$ for DVS Gesture), SAST reduces the surrogate-to-hard transfer gap by up to 92\% and 69\% respectively.
(iii) We evaluate under a hardware-aware inference simulation (weight quantization, fixed-point membrane, discrete leak, SynOps energy proxy).
(iv) We report corruption evaluations, training overhead, compute-matched controls, and explicit scope limitations for on-sensor deployment settings.

%------------------------------------------------------------------------
\section{Related Work}
\label{sec:related}

\paragraph{SNN training and the transfer gap.}
Surrogate-gradient methods~\cite{neftci_surrogate,zenke_superspike,bellec2018l2l,shrestha2018slayer,wu2018stbp} enable gradient-based SNN training but do not explicitly control the mismatch between the smooth surrogate and the hard threshold used at deployment.
ANN-to-SNN conversion~\cite{diehl2015fast,han2020rmp,bu2022optimal} and threshold calibration~\cite{li2021free,sengupta2019going} address a related but distinct gap---rate-coded ANN-to-SNN transfer---and typically require many time steps.
Quantization-aware SNN training~\cite{eshraghian2023lessons} targets hardware precision but not the surrogate-to-hard nonlinearity swap.

\paragraph{Sharpness-aware minimization.}
SAM~\cite{foret_sam} and ASAM~\cite{kwon2021asam} find flat minima by optimizing a neighborhood worst-case loss, with benefits extending to quantization robustness~\cite{nicholson2026sast}.
The core SAST idea was introduced by Nicholson in a 2026 arXiv preprint~\cite{nicholson2026sast}; this paper builds on that initial formulation with expanded theory and experiments.

%------------------------------------------------------------------------
\section{Method}
\label{sec:method}

\paragraph{Surrogate-forward LIF SNN.}
We define an $L$-layer LIF network unrolled for $T$ time steps.
Each layer $\ell$ has weight matrix $A^{(\ell)}$, bias $b^{(\ell)}$, threshold $\theta^{(\ell)}$, and leak $\alpha\in(0,1)$.
The membrane potential evolves as
\begin{equation}
u_t^{(\ell)} = \alpha\, u_{t\!-\!1}^{(\ell)} + A^{(\ell)} \tilde s_t^{(\ell\!-\!1)} + b^{(\ell)} - \theta^{(\ell)}\!\odot\tilde s_{t\!-\!1}^{(\ell)},
\label{eq:membrane}
\end{equation}
with surrogate spikes $\tilde s_t^{(\ell)}=\sigma(u_t^{(\ell)}-\theta^{(\ell)})$ and input convention $\tilde s_t^{(0)}=x_t$, where $\sigma$ is an admissible smooth surrogate ($\sigma\!\in\!C^2$). For theory, $B_1,B_2$ denote \emph{local} derivative bounds on the visited membrane-offset region $\mathcal U$: $B_1\defeq\sup_{z\in\mathcal U}|\sigma'(z)|$, $B_2\defeq\sup_{z\in\mathcal U}|\sigma''(z)|$. We use the arctan surrogate $\sigma(x)=\frac{1}{2}+\frac{1}{\pi}\arctan(kx)$ (with $k{=}25$ in experiments), whose global slope bound is $B_1^{\mathrm{glob}}=k/\pi$.
For clarity, we use this delayed-reset update order everywhere: compute $u_t^{(\ell)}$ with the reset term $\tilde s_{t-1}^{(\ell)}$, then compute spikes at the same step from $u_t^{(\ell)}$; therefore spikes emitted at time $t$ affect the reset at time $t\!+\!1$. This exact convention is used in both the theoretical model and all experiments.
The readout is $\tilde f_w(x_{1:T})=W_{\text{out}}(\frac{1}{T}\sum_t \tilde s_t^{(L)})+b_{\text{out}}$.

At \emph{on-sensor deployment}, $\sigma$ is replaced by the Heaviside step $H$---this is the only change.
Hard-spike evaluation uses the same trained weights, thresholds, leak, and reset rule, with all hidden states reset per sequence and no post-hoc calibration.

\begin{figure}[t]
\centering
\scriptsize
\begin{tikzpicture}[
>=Latex,
node distance=2.5mm and 2.8mm,
box/.style={draw, rounded corners=1pt, thick, align=center, minimum height=8.5mm, text width=0.27\linewidth, fill=blue!4},
deploy/.style={draw, rounded corners=1pt, thick, align=center, minimum height=6mm, text width=0.93\linewidth, fill=green!5}
]
\node[box] (s1) {1) Surrogate forward/backward\\on minibatch $B$\\to obtain $g$};
\node[box, right=of s1] (s2) {2) Ascent perturbation\\$\epsilon=\rho g/(\norm{g}_2+\delta)$};
\node[box, right=of s2] (s3) {3) Reset states\\and compute gradient\\at $w+\epsilon$ on $B'$};
\node[box, below=of s2] (s4) {4) Optimizer update\\using gradient at $w+\epsilon$};
\node[deploy, below=of s4] (deploy) {Deployment: replace surrogate $\sigma$ with hard step $H$; keep weights, thresholds, leak, and reset rule unchanged.};

\draw[->, thick] (s1) -- (s2);
\draw[->, thick] (s2) -- (s3);
\draw[->, thick] (s2) -- (s4);
\draw[->, thick] (s4) -- (deploy);
\end{tikzpicture}
\caption{SAST pipeline: two-pass SAM with state reset, then hard-spike deployment without post-hoc calibration.}
\label{fig:sast_pipeline}
\end{figure}
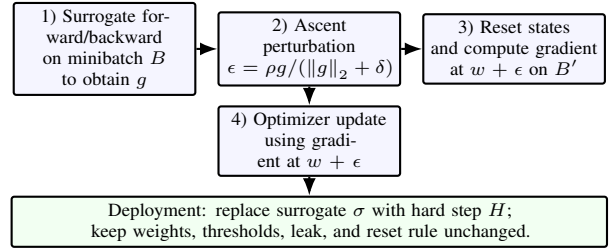

\paragraph{SAST algorithm.}
SAST applies SAM to the surrogate-forward empirical risk $\tilde L_S(w)=\frac{1}{n}\sum_i \ell(\tilde f_w(x^{(i)}_{1:T}),y^{(i)})$:
\begin{equation}
\tilde L_{\text{SAM}}(w)\defeq \max_{\norm{\epsilon}_2\le\rho}\tilde L_S(w+\epsilon).
\end{equation}
Each training step (i)~computes the surrogate loss and gradient $g$ on minibatch~$B$, (ii)~forms the ascent perturbation $\epsilon=\rho\,g/(\norm{g}_2+\delta)$, (iii)~resets all SNN states and computes the gradient at $w+\epsilon$ on an independent minibatch~$B'$, and (iv)~updates $w$ with the optimizer.
State resets between SAM passes prevent stale temporal state from confounding the perturbation.
\paragraph{Notation for constants.}
In the SAM ascent step, $\delta>0$ is a small fixed numerical stabilization constant added to $\norm{g}_2$ to avoid division by zero; it is a user-set implementation constant (assumed/fixed, not estimated from data).
Figure~\ref{fig:sast_pipeline} summarizes the end-to-end SAST training/deployment flow used in this paper.

%------------------------------------------------------------------------
\section{Theoretical Guarantees}
\label{sec:theory}

Under bounded inputs ($\norm{x_t}_2\!\le\!R_x$), bounded operator norms ($\norm{A^{(\ell)}}_2\!\le\!M_A$, $\norm{\theta^{(\ell)}}_\infty\!\le\!M_\theta$), and a one-step \emph{local} contraction condition $\gamma\defeq\alpha+M_\theta B_1<1$ (with $B_1$ defined on $\mathcal U$ above, not set to the global $k/\pi$ bound), we establish:

\paragraph{Notation for constants.}
$M_{\text{out}}$ denotes an assumed operator-norm bound for the readout map (for a linear readout, $M_{\text{out}}\defeq\norm{W_{\text{out}}}_2$). In Eq.~\eqref{eq:conv}, $\tilde L^\star$ denotes the optimal (infimum) surrogate objective value, and $\sigma_{\text{noise}}^2$ denotes the minibatch-gradient noise variance bound. Here $M_{\text{out}}$, $\tilde L^\star$, and $\sigma_{\text{noise}}^2$ are analysis constants (assumed/theoretical); empirical counterparts may be estimated from checkpoints but are not directly measured primary metrics.

\textbf{State stability \& input Lipschitz bound.}
Surrogate membrane potentials are uniformly bounded, and the readout satisfies
$\norm{\tilde f_w(x_{1:T})-\tilde f_w(x'_{1:T})}_2 \le L_x\norm{x_{1:T}-x'_{1:T}}_{2,2}$
with $L_x = M_{\text{out}}(B_1 M_A S_T(\gamma))^L/\!\sqrt{T}$, where $S_T(\gamma)=(1-\gamma^T)/(1-\gamma)$.
This bounds how much on-sensor input perturbations (e.g.\ event drops, noise) can affect predictions.

\textbf{Smoothness.}
The empirical surrogate objective $\tilde L_S$ is $\beta$-smooth, with $\beta$ depending on depth, temporal gain, and surrogate slope.

\textbf{First-order SAM view.}
Smoothness gives $\tilde L_{\text{SAM}}(w)\le \tilde L_S(w)+\rho\norm{\nabla\tilde L_S(w)}_2+\tfrac{\beta\rho^2}{2}$, so minimizing $\tilde L_{\text{SAM}}$ approximately penalizes gradient norm~\cite{foret_sam,wen2023samsharpness}.

\textbf{Convergence.}
With independent second minibatches and step size $\eta\le 1/(4\beta)$,
\begin{equation}
\frac{1}{K}\!\sum_{k=0}^{K\!-\!1}\!\E\norm{\nabla\tilde L_S(w_k)}_2^2
\!\le\!
\frac{4(\tilde L_S(w_0)\!-\!\tilde L^\star)}{\eta K}
\!+\! 3\beta^2\rho^2
\!+\! 2\eta\beta\sigma_{\text{noise}}^2.
\label{eq:conv}
\end{equation}
The SAM perturbation contributes an additive $O(\beta^2\rho^2)$ floor.

%------------------------------------------------------------------------
\section{Experiments}
\label{sec:experiments}

% =====================================================================
% EXPANDED: Setup with architecture, hyperparameters, contraction diag.
% =====================================================================

\paragraph{Setup.}
We evaluate on \textbf{N-MNIST}~\cite{nmnist} and \textbf{DVS Gesture}~\cite{amir2017lowpower}, two event-camera benchmarks that represent the data modality of on-sensor vision systems.

\paragraph{Architecture.}
We use a dataset-adaptive 3-layer fully-connected LIF architecture (\texttt{fc\_0p40m}) so each benchmark stays near the same memory budget.
For N-MNIST, the flattened input is $2312$ and the network is $2312\!\rightarrow\!168\!\rightarrow\!64\!\rightarrow\!10$ with $400{,}050$ learnable parameters and $232$ fixed thresholds ($400{,}282$ stored parameters total); for DVS Gesture, the flattened input is $4608$ and the network is $4608\!\rightarrow\!80\!\rightarrow\!336\!\rightarrow\!11$ with $399{,}643$ learnable parameters and $416$ fixed thresholds ($400{,}059$ stored parameters total).
Learnable counts include all weights and biases (the readout uses weights and bias only); fixed thresholds are reported separately as stored constants.
To stress-test architecture dependence, we use a small convolutional SNN (\texttt{conv\_0p42m}) with three $3{\times}3$ LIF-convolution blocks (channels $68/136/272$), global average pooling, and a linear readout, totaling $421{,}066$ (N-MNIST) and $421{,}339$ (DVS Gesture) stored parameters under the same temporal binning and optimizer schedule.

\paragraph{Training details.}
Event streams are temporally binned into $T{=}10$ frames, normalized to $[0,1]$.
We use Adam with learning rate $10^{-3}$, cosine annealing over 200 epochs (N-MNIST) / 300 epochs (DVS Gesture), batch size 128, leak $\alpha{=}0.5$, threshold $\theta{=}1.0$, and arctan surrogate with slope $k{=}25$.
For this surrogate, the global derivative maximum is $k/\pi\approx7.96$; the contraction diagnostics below therefore use the empirical \emph{local} slope on visited states.
All swap-only hard-spike numbers replace only the surrogate nonlinearity, with no recalibration or threshold tuning.
For new datasets, we recommend a short sweep over $\rho\in\{0.10,0.20,0.30,0.40,0.50\}$; in our runs, the best swap-only hard-spike transfer occurred at $\rho\!=\!0.30$ (N-MNIST) and $\rho\!=\!0.40$ (DVS Gesture).
\paragraph{Why grid search for $\rho$.}
We tune $\rho$ with a short sweep because swap-only hard-spike accuracy and $\Delta_{\text{transfer}}$ are non-monotonic in $\rho$.
Results are averaged over 5 seeds.

\paragraph{Contraction diagnostic.}
Section~\ref{sec:theory} assumes $\gamma := \alpha + M_\theta B_1 < 1$ with local $B_1=\sup_{z\in\mathcal U}|\sigma'(z)|$.
Per checkpoint, we report an empirical proxy
$\hat\gamma \defeq \alpha + \hat M_\theta\hat B_1$, where $\hat M_\theta\defeq\max_\ell\|\theta^{(\ell)}\|_\infty$ and $\hat B_1\defeq\max_{\ell,t,n}|\sigma'(u_t^{(\ell,n)}-\theta^{(\ell)})|$, computed on one full validation pass (max over layers, timesteps, and samples).
In our setup thresholds are fixed at $\theta{=}1.0$, so $\hat M_\theta{=}1$ and $\hat\gamma=\alpha+\hat B_1$. Consistent with the architecture paragraph above, fixed thresholds are not learnable and are reported separately from learnable parameter counts.
Across both datasets, measured $\hat\gamma$ values satisfied $\hat\gamma<1$, and SAST yielded tighter contraction than baseline.

\paragraph{Primary metric: transfer gap.}
We define $\Delta_{\text{transfer}} = \text{Acc}_{\text{sur}} - \text{Acc}_{\text{hard}}$.
For on-sensor deployment, minimizing $\Delta_{\text{transfer}}$ while preserving surrogate accuracy is the central goal: a model is only as useful as its swap-only hard-spike accuracy before hardware-aware constraints are applied.

\begin{table}[t]
\caption{Main results: surrogate-forward and swap-only hard-spike accuracy, plus transfer gap. Best values are in \textbf{bold}.}
\label{tab:main}
\centering
\footnotesize
\setlength{\tabcolsep}{2.4pt}
\resizebox{\columnwidth}{!}{%
\begin{tabular}{llccc}
\toprule
Dataset & Method & Surrogate forward & Swap-only hard-spike & $\Delta_{\text{transfer}}$ \\
\midrule
N-MNIST & Baseline surrogate training & .9606$\pm$.0033 & .6572$\pm$.0974 & .3034 \\
N-MNIST & SAST $\rho\!=\!0.10$ & \textbf{.9786$\pm$.0009} & .8335$\pm$.0894 & .1451 \\
N-MNIST & SAST $\rho\!=\!0.30$ & .9721$\pm$.0012 & \textbf{.9473$\pm$.0462} & \textbf{.0248} \\
\midrule
DVS Gest.\ & Baseline surrogate training & .7502$\pm$.0142 & .3182$\pm$.0732 & .4320 \\
DVS Gest.\ & SAST $\rho\!=\!0.20$ & \textbf{.8087$\pm$.0043} & .5957$\pm$.0116 & .2130 \\
DVS Gest.\ & SAST $\rho\!=\!0.40$ & .7685$\pm$.0151 & \textbf{.6327$\pm$.0116} & \textbf{.1358} \\
\bottomrule
\end{tabular}%
}
\end{table}

\begin{table}[t]
\caption{Added controls under swap-only hard-spike evaluation: Conv-SNN architecture ablation. Best values are in \textbf{bold}.}
\label{tab:added_controls}
\centering
\footnotesize
\setlength{\tabcolsep}{2.2pt}
\resizebox{\columnwidth}{!}{%
\begin{tabular}{lllccc}
\toprule
Dataset & Arch./control & Method & Surrogate forward & Swap-only hard-spike & $\Delta_{\text{transfer}}$ \\
\midrule
N-MNIST & Conv-SNN (${\sim}0.42$M) & Baseline surrogate training & .9868$\pm$.0009 & .9091$\pm$.0153 & .0777 \\
N-MNIST & Conv-SNN (${\sim}0.42$M) & SAST (best $\rho$) & \textbf{.9877$\pm$.0003} & \textbf{.9733$\pm$.0032} & .0144 \\
\midrule
DVS Gest. & Conv-SNN (${\sim}0.42$M) & Baseline surrogate training & \textbf{.7438$\pm$.0044} & .4630$\pm$.0075 & .2808 \\
DVS Gest. & Conv-SNN (${\sim}0.42$M) & SAST (best $\rho$) & .7284$\pm$.0115 & \textbf{.6204$\pm$.0151} & .1080 \\
\bottomrule
\end{tabular}%
}
\end{table}

\begin{figure}[t]
\centering
\includegraphics[width=\linewidth]{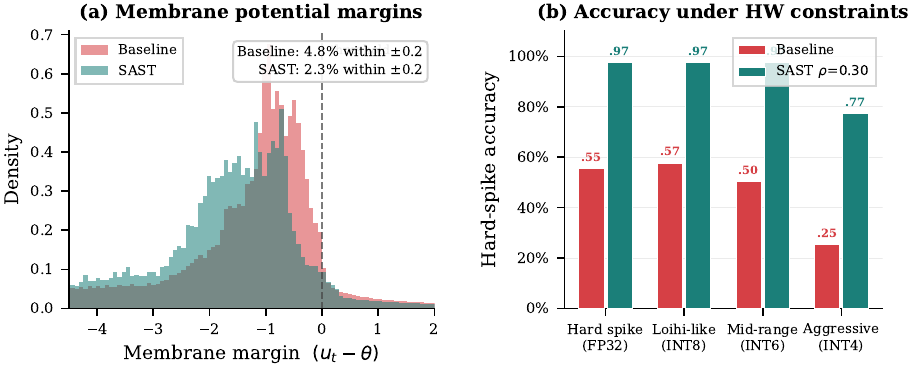}
\caption{(a) Swap-only hard-spike membrane margins $(u_t\!-\!\theta)$ on N-MNIST. Baseline mass within $\pm 0.2$: 4.8\%; SAST: 2.3\%. (b) Hardware-aware hard-spike accuracy under hardware constraints; on N-MNIST at INT8/Q8.8, SAST is about 97\% vs.\ about 48\% for baseline.}
\label{fig:margins_hw}
\end{figure}

\paragraph{Results.}
Table~\ref{tab:main} shows that SAST strongly reduces transfer gap for the fully-connected architecture while improving swap-only hard-spike accuracy.
On N-MNIST, $\Delta_{\text{transfer}}$ falls from 0.303 to 0.025 (92\% relative reduction) at $\rho\!=\!0.30$, with swap-only hard-spike accuracy rising from 65.7\% to 94.7\%.
On DVS Gesture, the gap drops from 0.432 to 0.136 (+31.5 pp swap-only hard-spike accuracy).
Surrogate-forward accuracy remains high or improves slightly at moderate~$\rho$, confirming SAM does not sacrifice the training signal.
At lower $\rho$ (e.g., $\rho\!=\!0.10$), seed-to-seed variance is larger, suggesting a minimum perturbation strength is needed for flat, transfer-friendly regions.
Figure~\ref{fig:margins_hw}(a) reveals the mechanism: SAST halves the fraction of membrane potentials in the ambiguous zone near threshold.

\paragraph{Corruption evaluation and training overhead.}
Under random event-drop corruption ($p\!\in\!\{0.0,0.1,0.2,0.3,0.4\}$), SAST is at or above baseline across the tested levels (equal at $p\!=\!0.00$). In this test, the largest observed separation is at $p\!=\!0.40$, where SAST reaches 96.1\% versus 90.8\% for baseline (${+}5.3$ percentage points), with a 3.1-point difference at $p\!=\!0.30$ (97.3\% vs.\ 94.2\%). From clean input to $p\!=\!0.40$, the accuracy drop is 2.1 points for SAST versus 7.4 points for baseline.

SAM doubles per-step gradient cost (${\sim}2.1\times$ wall-clock on N-MNIST, ${\sim}1.8\times$ on DVS Gesture); with sequential loading of $B$ and $B'$ we observe no increase in peak memory.

\paragraph{Compute-matched baseline.}
Since SAM increases per-step cost, we compare wall-clock compute-matched budgets and report best-achieved epochs.
For each method, dataset, and seed, checkpoint selection is based on the highest \emph{validation} swap-only hard-spike accuracy on a held-out validation split; we then report test metrics once at that selected epoch (the test set is not used for model selection).
Table~\ref{tab:compute_matched} summarizes this comparison.
On N-MNIST, the compute-matched baseline (100 epochs) reaches 65.7\% swap-only hard-spike accuracy with $\Delta_{\text{transfer}}=0.303$, while SAST (48 epochs) reaches 93.9\% with $\Delta_{\text{transfer}}=0.036$.
On DVS Gesture, the compute-matched baseline (400 epochs) reaches 28.0\% swap-only hard-spike accuracy with $\Delta_{\text{transfer}}=0.504$, while SAST (226 epochs) reaches 57.8\% with $\Delta_{\text{transfer}}=0.125$.
This comparison suggests that extra baseline budget alone does not close the transfer gap.

\begin{table}[t]
\caption{Compute-matched comparison under swap-only hard-spike evaluation (wall-clock matched); SAST uses per-dataset best $\rho$.}
\label{tab:compute_matched}
\centering
\footnotesize
\setlength{\tabcolsep}{2.6pt}
\resizebox{\columnwidth}{!}{%
\begin{tabular}{llccc}
\toprule
Dataset & Method & Epochs & Swap-only hard-spike & $\Delta_{\text{transfer}}$ \\
\midrule
N-MNIST & Baseline (compute-matched) & 100 & .657$\pm$.097 & .303 \\
N-MNIST & SAST (best $\rho$) & 48 & .939$\pm$.022 & .036 \\
\midrule
DVS Gest. & Baseline (compute-matched) & 400 & .280$\pm$.070 & .504 \\
DVS Gest. & SAST (best $\rho$) & 226 & .578$\pm$.038 & .125 \\
\bottomrule
\end{tabular}%
}
\end{table}

\subsection{Hardware-aware inference simulation}
\label{subsec:hw_sim}

To avoid notation drift, Table~\ref{tab:main} reports \emph{swap-only hard-spike} inference (only surrogate $\sigma\to H$ replacement).
Here we report independent runs under a hardware-aware pipeline (quantized weights, fixed-point membrane, and discrete leak) with hardware-aware hard-spike inference and reset-by-subtraction.
Table~\ref{tab:hw_sim} summarizes representative operating points: Loihi-like INT8/Q8.8 and aggressive INT4/Q4.4, plus SynOps as an activity-dependent energy proxy.
For one sample-sequence $x_{1:T}$, we define
\[
\mathrm{SynOps}(x_{1:T})\defeq \sum_{t=1}^{T}\sum_{\ell}\sum_{n} z_{t,n}^{(\ell)}F_{t,n}^{(\ell)},
\]
where $z_{t,n}^{(\ell)}\in\{0,1\}$ is presynaptic spike activity and $F_{t,n}^{(\ell)}$ is the implemented fan-out (number of downstream synaptic accumulations triggered by that spike, including boundary effects for convolution).
We report $k\mathrm{SynOps}=10^{-3}\,\E_{x\sim\mathcal V}[\mathrm{SynOps}(x_{1:T})]$: thousands of synaptic accumulations per sample-sequence, using the same $T{=}10$ temporal binning as the main experiments (not per second).

\begin{table}[t]
\caption{Hardware-aware hard-spike summary (independent runs). kSynOps: synaptic accumulations per sample-sequence ($\times 10^3$, $T{=}10$), and $r_{\text{ops}}\defeq \mathrm{kSynOps}_{\text{SAST}}/\mathrm{kSynOps}_{\text{Baseline}}$.}
\label{tab:hw_sim}
\centering
\footnotesize
\setlength{\tabcolsep}{2.8pt}
\resizebox{\columnwidth}{!}{%
\begin{tabular}{llccccc}
\toprule
Dataset & Profile & Baseline acc. & SAST acc. & Baseline kSynOps & SAST kSynOps & $r_{\text{ops}}$ \\
\midrule
N-MNIST & Loihi-like (INT8, Q8.8) & 0.476 & \textbf{0.969} & 1734.0 & 1315.0 & 0.758 \\
N-MNIST & Aggressive (INT4, Q4.4) & 0.432 & \textbf{0.810} & 1666.0 & 1346.0 & 0.808 \\
DVS Gest. & Loihi-like (INT8, Q8.8) & 0.253 & \textbf{0.476} & 86221.3 & 4323.5 & 0.050 \\
DVS Gest. & Aggressive (INT4, Q4.4) & 0.260 & \textbf{0.438} & 82317.0 & 4145.6 & 0.050 \\
\bottomrule
\end{tabular}%
}
\end{table}

Table~\ref{tab:hw_sim} shows that SAST's advantage persists for hardware-aware hard-spike inference under hardware constraints.
At Loihi-like precision, SAST reaches 96.9\% vs.\ 47.6\% on N-MNIST and 47.6\% vs.\ 25.3\% on DVS Gesture; at INT4, it reaches 81.0\% vs.\ 43.2\% on N-MNIST and 43.8\% vs.\ 26.0\% on DVS Gesture.
SAST also reduces SynOps, including 1734k to 1315k on N-MNIST (INT8) and 86221k to 4323k on DVS Gesture (INT8).
These trends are consistent with sharpness-aware training pushing membrane potentials away from the decision boundary, producing spike patterns that are more robust to both the hard-spike swap and low-precision arithmetic.
Figure~\ref{fig:margins_hw}(b) summarizes these results visually.

%------------------------------------------------------------------------
\section{Discussion and On-Sensor Relevance}
\label{sec:discussion}

Under the tested settings, SAST reduces the surrogate-to-hard transfer gap for on-sensor deployment without post-hoc calibration or quantization-aware retraining.

{\small
\bibliographystyle{IEEEtranN}
\bibliography{references}
}

\end{document}